\documentclass{article}

\usepackage{amsmath, amssymb, amsthm}
\usepackage{graphicx}
\usepackage{hyperref}
\usepackage{algorithm}
\usepackage{algorithmic}
\usepackage{booktabs}
\usepackage{datatool}
\usepackage{fullpage}
\usepackage{natbib}
\usepackage{color}
\usepackage{listings}
\usepackage{float}
\usepackage{caption}

\usepackage{tikz}
\usetikzlibrary{positioning, shapes.geometric, arrows, calc, decorations.pathreplacing}
\usepackage{pgfplots}
\pgfplotsset{compat=1.17}

\title{Loop Neural Networks for Parameter Sharing}

\author{%
\begin{minipage}[t]{0.45\textwidth}
\centering
Kei-Sing Ng\textsuperscript{1} \\
\texttt{}
\end{minipage}%
\hfill
\begin{minipage}[t]{0.45\textwidth}
\centering
Qingchen Wang\textsuperscript{1} \\
\texttt{}
\end{minipage}
}

\date{}

\begin{document}

\maketitle

\begin{abstract}
The success of large-scale language models like GPT can be attributed to their ability to efficiently predict the next token in a sequence. However, these models rely on constant computational effort regardless of the complexity of the token they are predicting, lacking the capacity for iterative refinement. In this paper, we introduce a novel loop neural network, which achieves better performance by utilizing longer computational time without increasing the model size. Our approach revisits the input multiple times, refining the prediction by iteratively looping over a subset of the model with residual connections. We demonstrate the effectiveness of this method through experiments comparing versions of GPT-2 with our loop neural networks, showing improved performance in language modeling tasks while maintaining similar parameter counts. Importantly, these improvements are achieved without the need for extra training data.
\end{abstract}

\section{Introduction}

The transformer architecture has revolutionized natural language processing, enabling models like GPT to predict the next token in a sequence efficiently \citep{vaswani2017attention, radford2019language}. Despite their success, these models perform a one-pass projection of all previous tokens to predict the next token, which limits their capacity for \textit{iterative refinement}. Specifically, they rely on constant computational effort regardless of the complexity or ambiguity of the token being predicted.

In this paper, we introduce a novel loop neural network, which revisits the input multiple times, refining the prediction by iteratively looping over a subset of the model with residual connections. Our main contribution is improving transformer performance with longer inference times, using a novel loop architecture with residual prediction. This approach works for large neural networks without needing extra training data, effectively extending the model's approximation capacity.

Our contributions are significant for the following reasons:

\begin{itemize}
\item \textbf{Novel Architecture:} We introduce a loop neural network mechanism that enhances model performance without increasing parameter count, making it accessible for researchers with limited computational resources.
\item \textbf{Efficiency:} By utilizing longer inference times, our model achieves better performance without the need for extra training data, contrasting with approaches that rely on extensive data augmentation.
\item \textbf{Scalability:} Our method is applicable to large-scale neural networks, demonstrating effectiveness on models comparable to GPT-2.
\end{itemize}

We demonstrate the effectiveness of this method through experiments comparing versions of GPT-2 with our loop neural networks, showing that our loop neural network with 6 transformer layers achieves a validation loss of 3.11 on the OpenWebText dataset \citep{Gokaslan2019OpenWebText}, comparable to the GPT2 model's loss of 3.12 with 12 transformer layers . Similarly, our loop neural network with 4 layers and 12 loops achieves a validation loss of 3.15, only less than 1\% increase compared to the GPT2-124M model. These results emphasize the benefit of utilizing longer inference times to improve performance without increasing model size or requiring additional data.

\section{Background and Related Work}

Traditional neural networks, including transformers, map a sequence of inputs to a prediction in a single forward pass, processing inputs through multiple layers to refine internal representations. The transformer architecture relies on self-attention mechanisms and feed-forward networks to capture dependencies in the data \citep{vaswani2017attention}.

Residual connections, introduced by \citet{he2016deep}, have been instrumental in enabling the training of very deep networks by alleviating the vanishing gradient problem. Our loop neural network leverages this concept by focusing on refining the hidden state through iterative loops, allowing the model to stabilize training and accelerate convergence without increasing its size significantly.

\subsection{Parameter Sharing and Adaptive Computation}

Several prior works have explored parameter sharing and adaptive computation in neural networks.

\subsubsection{Universal Transformers}

\citet{dehghani2018universal} introduced the Universal Transformer, which applies the transformer layers recurrently to capture both short-term and long-term dependencies. Their model iteratively refines representations by looping over the entire transformer block. However, their experiments were primarily conducted on smaller models and datasets, such as the WMT English-German translation task, and did not explore large-scale language models like GPT-2. Additionally, their design does not incorporate a predictive residual mechanism as we do, which is crucial for stabilizing training in large-scale models.

\subsubsection{Adaptive Computation Time Models}

\citet{graves2016adaptive} proposed Adaptive Computation Time (ACT) for recurrent neural networks, allowing models to dynamically decide the number of computational steps per input. While ACT introduces adaptive computation, it was mainly applied to simple RNN architectures and tested on small-scale tasks, without leveraging the transformer architecture or large-scale pretraining. Moreover, ACT does not employ a predictive residual design, which differentiates our approach.

\subsubsection{Depth-Adaptive Transformers}

\citet{elbayad2019depth} presented Depth-Adaptive Transformers that adjust the depth of the network based on the input. Their method enables dynamic inference by selecting the number of layers to apply per input sequence. However, their approach focuses on varying depth during inference rather than iterative refinement during training, and their experiments were conducted on machine translation tasks with relatively small models. Importantly, their model lacks the predictive residual design present in our loop architecture.

\subsubsection{Parameter Sharing Across Layers}

Parameter sharing across layers has been explored as a way to reduce the number of parameters in transformer models while maintaining performance. \citet{TakaseKiyono2023} investigated the effects of parameter sharing in transformers, showing that sharing parameters across layers can be effective under certain conditions. In our work, we employ a further idea by looping over layers with shared parameters with residual and gate design, effectively increasing computational depth without adding new parameters.

\section{Methodology}

We propose a novel architecture called the loop neural network, which refines the hidden state through iterative loops over a subset of transformer blocks with residual connections and gating mechanisms. This iterative process allows the model to revisit and enhance its internal representations multiple times, effectively increasing its approximation capacity without adding more parameters.

\subsection{General Loop Structure with Residual Design}

In our model, the hidden state is updated iteratively using an approach inspired by series expansions in numerical methods. Let $x^{(n)} \in \mathbb{R}^d$ denote the hidden state after the $n$-th iteration, where $x^{(0)}$ is the initial hidden state obtained from the embedding layer or the output of the previous layer.

The iterative update rule is defined as:

\begin{equation} \label{eq}
x^{(n)} = x^{(n-1)} + a_n \odot f_{\theta}\left( x^{(n-1)} \right),
\end{equation}

where:

\begin{itemize}
\item $x^{(n-1)} \in \mathbb{R}^d$ is the hidden state from the previous iteration.
\item $f_{\theta}\left( x^{(n-1)} \right) \in \mathbb{R}^d$ is the output of the transformer block(s) parameterized by $\theta$.
\item $a_n \in \mathbb{R}^d$ is a gating coefficient vector at iteration $n$, applied element-wise (denoted by $\odot$). The gating coefficients are learned parameters.
\item $a_0 = \mathbf{1}$ (a vector of ones).
\end{itemize}

By unrolling the iterative process, the hidden state after $N$ iterations can be expressed as:

\begin{align} \label{eq}
x^{(N)} = a_0 \odot x^{(0)} + \sum_{k=1}^{N} a_k \odot f_{\theta}\left( x^{(k-1)} \right).
\end{align}

Alternatively, we can represent the hidden state accumulation explicitly:

\begin{equation} \label{eq}
x^{(N)} = a_0 \odot x^{(0)} + a_1 \odot f_{\theta}\left( x^{(0)} \right) + a_2 \odot f_{\theta}\left( x^{(1)} \right) + \dots + a_N \odot f_{\theta}\left( x^{(N-1)} \right).
\end{equation}

Since $a_0 = \mathbf{1}$ and $f_{\theta}\left( x^{(k-1)} \right)$ represents the correction at each iteration, the model effectively accumulates the contributions from all iterations to refine the hidden state.

The iterative refinement process allows the model to capture complex patterns and dependencies by progressively updating the hidden state. The use of residual connections helps to alleviate issues such as vanishing gradients and facilitates the flow of information across iterations.

\subsection{Looping Strategies}

Our approach applies the loop in two ways: looping over all layers multiple times, and looping within each layer. This strategy effectively increases the computational depth of the model without increasing the parameter count, similar to techniques explored in parameter sharing across transformer layers.

\subsection{Illustration of the Loop Neural Network}

\begin{figure}[H]
\centering
\begin{tikzpicture}[
node distance=0.8cm and 1.0cm,
block/.style={rectangle, draw, fill=blue!10, rounded corners, minimum width=8em, minimum height=2em},
gate/.style={rectangle, draw, fill=yellow!20, rounded corners, minimum width=6em, minimum height=1.5em},
sum/.style={circle, draw, inner sep=0pt, minimum size=1.2em, fill=gray!20},
mul/.style={circle, draw, inner sep=0pt, minimum size=1.2em, fill=gray!20},
embedding/.style={rectangle, draw, fill=green!10, rounded corners, minimum width=10em, minimum height=2em},
output/.style={rectangle, draw, fill=red!10, rounded corners, minimum width=10em, minimum height=2em},
arrow/.style={->, >=stealth, thick},
on grid,
every node/.style={inner sep=2pt},
]

\node[embedding] (emb) at (0,0) {Embedding Layer};

\node[below=of emb] (x_prev) {$x^{(n-1)}$};

\node[block, below=of x_prev] (block1) {Transformer Block 1};
\node[block, below=of block1] (block2) {Transformer Block 2};
\node[block, below=of block2] (block3) {Transformer Block 3};

\node[gate, below=of block3] (a_n) {Gating Coefficient $a_n$};

\node[mul, right=3cm of a_n] (mul) {$\odot$};

\node[sum, right=3cm of mul] (sum) {$+$};

\node[right=3cm of sum] (x_n) {$x^{(n)}$};

\node[output, below=of x_n] (output) {Output Layer};

\draw[arrow] (emb) -- (x_prev);
\draw[arrow] (x_prev) -- (block1);
\draw[arrow] (block1) -- (block2);
\draw[arrow] (block2) -- (block3);
\draw[arrow] (block3) -- (a_n);

\draw[arrow] (a_n.east) -| (mul.south);

\draw[arrow] (x_prev.east) -| ($(sum.north)+(0,0.2cm)$) -- (sum.north);

\draw[arrow] (mul) -- (sum);
\draw[arrow] (sum) -- (x_n);

\draw[arrow] (x_n) -- (output);

\draw[arrow] (x_n.west) -| node[pos=0.25, above] {Repeat for $n = 1$ to $N$} (x_prev.east);

\end{tikzpicture}
\caption{Computational Graph with Multiple Transformer Blocks}
\label{fig:computational_graph_blocks}
\end{figure}
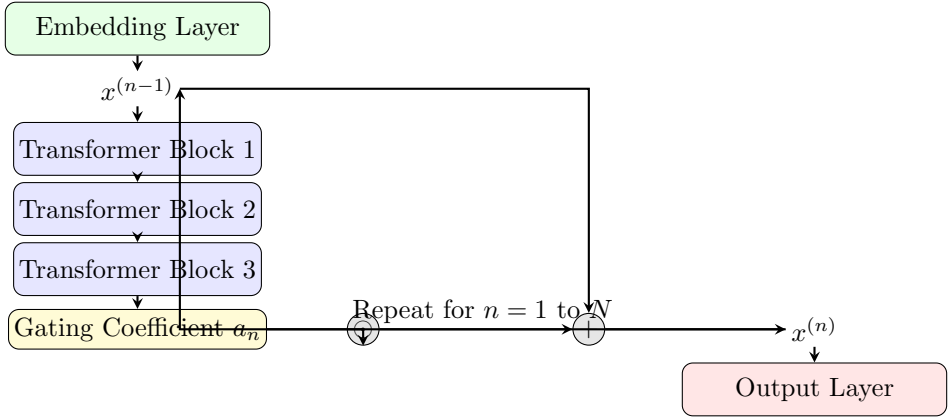

\subsection{Comparison with Existing Methods}

Our approach differs from existing methods by explicitly focusing on predicting the residual at each iteration and adding it back to the hidden state. Unlike Universal Transformers that loop over the entire transformer block without residual prediction, our method leverages residual connections within the looping mechanism to stabilize training and enhance convergence, especially in large-scale models.

\section{Experiments}

We conducted experiments to evaluate the effectiveness of our loop neural networks in improving transformer performance using longer inference times without increasing model size. Our experiments focus on comparing our models with standard GPT-2 models of similar or larger sizes on the OpenWebText dataset.

\subsection{Experimental Setup}

In our first experiment, we compared loop neural networks with different configurations to the standard GPT-2 models. Specifically, we evaluated:

\begin{itemize}
\item \textbf{GPT2-124M}: The standard GPT-2 model with 124 million parameters and 12 layers.
\item \textbf{GPT2-81M-LOOP}: Our loop neural network with 81 million parameters, 6 layers, and 6 loops.
\item \textbf{GPT2-67M-LOOP}: Our loop neural network with 67 million parameters, 4 layers, and 12 loops.
\end{itemize}

All other structural and design elements of the model, such as the transformer block configurations and parameter initialization, remain identical to the baseline GPT-2 models, with the only modifications being the loop residual and gating mechanisms. The GPT2-124M model serves as our baseline. Our GPT2-81M-LOOP model employs 6 loops over 6 transformer layers, effectively simulating deeper computation without increasing the number of parameters. The GPT2-67M-LOOP model uses 6 loops over 4 transformer layers and 2 loops over each layer. This model demonstrates that even with fewer layers and parameters, we can achieve comparable performance by increasing the number of loops.

In our second experiment, we focused on smaller-scale models to illustrate the effectiveness of our method at different scales. We compared a loop neural network with 45 million parameters (\textbf{GPT2-45M-LOOP}) to a GPT2-45M model of the same size. The GPT2-45M model consists of a single transformer block layer, performing one-pass prediction, and serves as the baseline. Our GPT2-45M-LOOP model loops twice over a single transformer block to predict the residual. The objective here is to illustrate that even at smaller scales, our method can improve performance over the baseline without increasing model size by leveraging iterative refinement.

We selected these experimental settings to evaluate the efficacy of our loop neural network across different model sizes and to show that iterative refinement can compensate for a reduced number of parameters or layers. By keeping the total number of parameters constant while adjusting the depth via looping, we aim to isolate the effect of iterative refinement on model performance.

In all experiments, we used the same training data and hyperparameters where applicable to ensure a fair comparison. The models were trained on the OpenWebText dataset, which provides a diverse range of textual data suitable for language modeling tasks. We measured the training epoch times for our models and the standard GPT-2 models to assess the computational overhead introduced by our method. The training epoch times were 150ms for GPT2-45M, 177ms for our GPT2-45M-LOOP, and 1,377ms for GPT2-81M-LOOP.

\subsection{Evaluation Metrics}

We evaluate the models based on the cross-entropy loss on the training dataset (\textbf{Training Loss}) and the cross-entropy loss on a held-out validation dataset (\textbf{Validation Loss}). These metrics provide a standard measure of model performance in language modeling tasks, allowing us to assess the effectiveness of our approach in reducing prediction error.

\section{Results}

Our experiments demonstrate that the loop neural networks significantly improve performance over standard transformer models by leveraging longer inference times without increasing model size.

\subsection{First Experiment: Loop Neural Networks vs. GPT2-124M}

Our GPT2-81M-LOOP model achieved a validation loss of 3.11 on the OpenWebText dataset, comparable to the GPT2-124M model's loss of 3.12. Similarly, the GPT2-67M-LOOP model, with only 67 million parameters and 4 layers, achieved a validation loss of 3.15, only less than 1\% increase compared to the GPT2-124M model. These results are notable because our models use significantly fewer parameters and layers compared to the GPT2-124M model. The improvement indicates that iterative refinement through our looping mechanism effectively enhances the model's approximation capacity, allowing it to match or exceed the performance of larger models.

We do not have the training loss data for the GPT2-124M model since we evaluated it using the official release model parameters. In our graphs, we represent the GPT2-124M model's validation loss as a dotted line to indicate that it is a reference point rather than a directly comparable training curve.

\begin{table}[H]
\centering
\caption{Performance Metrics Emphasizing Layers and Loops for First Experiment}
\label{tab:first_experiment}
\begin{tabular}{cccccc}
\toprule
\textbf{Model} & \textbf{Parameters} & \textbf{Layers} & \textbf{Loops} & \textbf{Train Loss} & \textbf{Validation Loss} \\
\midrule
GPT2-124M & 124M & 12 & 1 & -- & 3.12 \\
GPT2-81M-LOOP & 81M & 6 & 6 & 3.13 & \textbf{3.11} \\
GPT2-67M-LOOP & 67M & 4 & 12 & 3.15 & 3.15 \\
\bottomrule
\end{tabular}
\end{table}

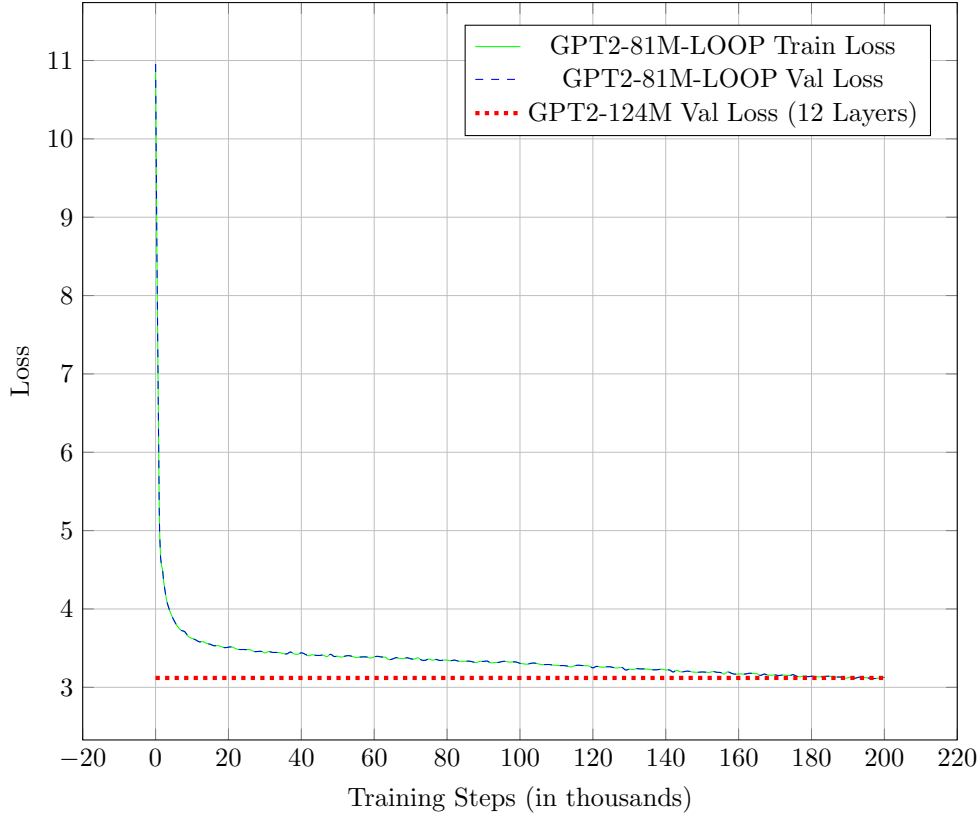
\begin{figure}[H]
\centering
\begin{tikzpicture}
\begin{axis}[
    width=0.8\linewidth,
    xlabel={Training Steps (in thousands)},
    ylabel={Loss},
    legend pos=north east,
    grid=both,
    grid style={line width=.1pt, draw=gray!10},
    major grid style={line width=.2pt,draw=gray!50},
]

\addplot[smooth, green] plot coordinates {
(0,10.9551362991333)
(1,5.28113842010498)
(2,4.46050119400024)
(3,4.12088298797607)
(4,3.96245861053467)
(5,3.85905456542969)
(6,3.77803516387939)
(7,3.7281699180603)
(8,3.71028757095337)
(9,3.64776539802551)
(10,3.62538599967957)
(11,3.60582256317139)
(12,3.58047342300415)
(13,3.5832986831665)
(14,3.56046175956726)
(15,3.54972815513611)
(16,3.53047728538513)
(17,3.53272342681885)
(18,3.52110147476196)
(19,3.50583076477051)
(20,3.51230335235596)
(21,3.51565837860107)
(22,3.49261808395386)
(23,3.48274660110474)
(24,3.48265981674194)
(25,3.48131418228149)
(26,3.47973418235779)
(27,3.45669746398926)
(28,3.45714974403381)
(29,3.45903849601746)
(30,3.4394953250885)
(31,3.45720648765564)
(32,3.44576478004456)
(33,3.44522523880005)
(34,3.44346714019775)
(35,3.43083882331848)
(36,3.42602515220642)
(37,3.45042014122009)
(38,3.42726016044617)
(39,3.42162132263184)
(40,3.44073247909546)
(41,3.42353820800781)
(42,3.40230846405029)
(43,3.41587972640991)
(44,3.40799188613892)
(45,3.40668725967407)
(46,3.41133737564087)
(47,3.39081454277039)
(48,3.42159914970398)
(49,3.39374876022339)
(50,3.39750051498413)
(51,3.38492441177368)
(52,3.39818811416626)
(53,3.40471196174622)
(54,3.39382457733154)
(55,3.38111758232117)
(56,3.38550543785095)
(57,3.38883304595947)
(58,3.38659596443176)
(59,3.3726761341095)
(60,3.38914084434509)
(61,3.39556050300598)
(62,3.38637328147888)
(63,3.38552498817444)
(64,3.35420846939087)
(65,3.36071085929871)
(66,3.37884449958801)
(67,3.36946797370911)
(68,3.36629509925842)
(69,3.37807464599609)
(70,3.36440801620483)
(71,3.3597686290741)
(72,3.37700724601746)
(73,3.34747862815857)
(74,3.34501767158508)
(75,3.3611843585968)
(76,3.34619474411011)
(77,3.35632610321045)
(78,3.34195137023926)
(79,3.34714961051941)
(80,3.3483304977417)
(81,3.33935332298279)
(82,3.34792852401733)
(83,3.33505654335022)
(84,3.33212041854858)
(85,3.33211660385132)
(86,3.33479309082031)
(87,3.32553720474243)
(88,3.31533718109131)
(89,3.33193397521973)
(90,3.32994389533997)
(91,3.33519291877747)
(92,3.31114959716797)
(93,3.31710338592529)
(94,3.3172779083252)
(95,3.32812476158142)
(96,3.33216404914856)
(97,3.32131481170654)
(98,3.32455205917358)
(99,3.3225781917572)
(100,3.30601859092712)
(101,3.2993369102478)
(102,3.29501438140869)
(103,3.3065447807312)
(104,3.30989813804626)
(105,3.29905366897583)
(106,3.29091620445251)
(107,3.29008841514587)
(108,3.29128551483154)
(109,3.28379201889038)
(110,3.28127002716064)
(111,3.27709603309631)
(112,3.27772641181946)
(113,3.27080726623535)
(114,3.25941824913025)
(115,3.27557134628296)
(116,3.28234529495239)
(117,3.27792000770569)
(118,3.27586245536804)
(119,3.27083158493042)
(120,3.24830436706543)
(121,3.26762080192566)
(122,3.25993347167969)
(123,3.25641846656799)
(124,3.2629919052124)
(125,3.26019525527954)
(126,3.24869847297668)
(127,3.24769997596741)
(128,3.253741979599)
(129,3.22109031677246)
(130,3.23170351982117)
(131,3.226154088974)
(132,3.23837351799011)
(133,3.23456239700317)
(134,3.23530149459839)
(135,3.23237538337708)
(136,3.22523260116577)
(137,3.22331047058105)
(138,3.22619009017944)
(139,3.23273110389709)
(140,3.21985983848572)
(141,3.22240877151489)
(142,3.19561648368835)
(143,3.2174026966095)
(144,3.19730281829834)
(145,3.19977688789368)
(146,3.20726633071899)
(147,3.19866847991943)
(148,3.19283819198608)
(149,3.19247364997864)
(150,3.19604802131653)
(151,3.19339466094971)
(152,3.18981504440308)
(153,3.20235657691956)
(154,3.19159078598022)
(155,3.18949866294861)
(156,3.16902160644531)
(157,3.19771814346313)
(158,3.17998194694519)
(159,3.16744136810303)
(160,3.17160487174988)
(161,3.16932320594788)
(162,3.17035531997681)
(163,3.18007588386536)
(164,3.17601704597473)
(165,3.17280220985413)
(166,3.15203523635864)
(167,3.16937470436096)
(168,3.1581597328186)
(169,3.15301513671875)
(170,3.15909457206726)
(171,3.14969158172607)
(172,3.15706086158752)
(173,3.15372252464294)
(174,3.1501898765564)
(175,3.16143798828125)
(176,3.13974738121033)
(177,3.13432598114014)
(178,3.14793515205383)
(179,3.13584208488464)
(180,3.14140009880066)
(181,3.139240026474)
(182,3.13279795646668)
(183,3.13629055023193)
(184,3.14004993438721)
(185,3.14193105697632)
(186,3.12688255310059)
(187,3.13093423843384)
(188,3.13094258308411)
(189,3.11580657958984)
(190,3.12074136734009)
(191,3.13199043273926)
(192,3.11605143547058)
(193,3.12983059883118)
(194,3.12584400177)
(195,3.12263512611389)
(196,3.11649203300476)
(197,3.11256217956543)
(198,3.11427855491638)
(199,3.12432670593262)
(200,3.12764406204224)

};
\addlegendentry{GPT2-81M-LOOP Train Loss}

\addplot[smooth, blue, dashed] plot coordinates {
(0,10.9551362991333)
(1,5.28113842010498)
(2,4.46050119400024)
(3,4.12088298797607)
(4,3.96245861053467)
(5,3.85905456542969)
(6,3.77803516387939)
(7,3.7281699180603)
(8,3.71028757095337)
(9,3.64776539802551)
(10,3.62538599967957)
(11,3.60582256317139)
(12,3.58047342300415)
(13,3.5832986831665)
(14,3.56046175956726)
(15,3.54972815513611)
(16,3.53047728538513)
(17,3.53272342681885)
(18,3.52110147476196)
(19,3.50583076477051)
(20,3.51230335235596)
(21,3.51565837860107)
(22,3.49261808395386)
(23,3.48274660110474)
(24,3.48265981674194)
(25,3.48131418228149)
(26,3.47973418235779)
(27,3.45669746398926)
(28,3.45714974403381)
(29,3.45903849601746)
(30,3.4394953250885)
(31,3.45720648765564)
(32,3.44576478004456)
(33,3.44522523880005)
(34,3.44346714019775)
(35,3.43083882331848)
(36,3.42602515220642)
(37,3.45042014122009)
(38,3.42726016044617)
(39,3.42162132263184)
(40,3.44073247909546)
(41,3.42353820800781)
(42,3.40230846405029)
(43,3.41587972640991)
(44,3.40799188613892)
(45,3.40668725967407)
(46,3.41133737564087)
(47,3.39081454277039)
(48,3.42159914970398)
(49,3.39374876022339)
(50,3.39750051498413)
(51,3.38492441177368)
(52,3.39818811416626)
(53,3.40471196174622)
(54,3.39382457733154)
(55,3.38111758232117)
(56,3.38550543785095)
(57,3.38883304595947)
(58,3.38659596443176)
(59,3.3726761341095)
(60,3.38914084434509)
(61,3.39556050300598)
(62,3.38637328147888)
(63,3.38552498817444)
(64,3.35420846939087)
(65,3.36071085929871)
(66,3.37884449958801)
(67,3.36946797370911)
(68,3.36629509925842)
(69,3.37807464599609)
(70,3.36440801620483)
(71,3.3597686290741)
(72,3.37700724601746)
(73,3.34747862815857)
(74,3.34501767158508)
(75,3.3611843585968)
(76,3.34619474411011)
(77,3.35632610321045)
(78,3.34195137023926)
(79,3.34714961051941)
(80,3.3483304977417)
(81,3.33935332298279)
(82,3.34792852401733)
(83,3.33505654335022)
(84,3.33212041854858)
(85,3.33211660385132)
(86,3.33479309082031)
(87,3.32553720474243)
(88,3.31533718109131)
(89,3.33193397521973)
(90,3.32994389533997)
(91,3.33519291877747)
(92,3.31114959716797)
(93,3.31710338592529)
(94,3.3172779083252)
(95,3.32812476158142)
(96,3.33216404914856)
(97,3.32131481170654)
(98,3.32455205917358)
(99,3.3225781917572)
(100,3.30601859092712)
(101,3.2993369102478)
(102,3.29501438140869)
(103,3.3065447807312)
(104,3.30989813804626)
(105,3.29905366897583)
(106,3.29091620445251)
(107,3.29008841514587)
(108,3.29128551483154)
(109,3.28379201889038)
(110,3.28127002716064)
(111,3.27709603309631)
(112,3.27772641181946)
(113,3.27080726623535)
(114,3.25941824913025)
(115,3.27557134628296)
(116,3.28234529495239)
(117,3.27792000770569)
(118,3.27586245536804)
(119,3.27083158493042)
(120,3.24830436706543)
(121,3.26762080192566)
(122,3.25993347167969)
(123,3.25641846656799)
(124,3.2629919052124)
(125,3.26019525527954)
(126,3.24869847297668)
(127,3.24769997596741)
(128,3.253741979599)
(129,3.22109031677246)
(130,3.23170351982117)
(131,3.226154088974)
(132,3.23837351799011)
(133,3.23456239700317)
(134,3.23530149459839)
(135,3.23237538337708)
(136,3.22523260116577)
(137,3.22331047058105)
(138,3.22619009017944)
(139,3.23273110389709)
(140,3.21985983848572)
(141,3.22240877151489)
(142,3.19561648368835)
(143,3.2174026966095)
(144,3.19730281829834)
(145,3.19977688789368)
(146,3.20726633071899)
(147,3.19866847991943)
(148,3.19283819198608)
(149,3.19247364997864)
(150,3.19604802131653)
(151,3.19339466094971)
(152,3.18981504440308)
(153,3.20235657691956)
(154,3.19159078598022)
(155,3.18949866294861)
(156,3.16902160644531)
(157,3.19771814346313)
(158,3.17998194694519)
(159,3.16744136810303)
(160,3.17160487174988)
(161,3.16932320594788)
(162,3.17035531997681)
(163,3.18007588386536)
(164,3.17601704597473)
(165,3.17280220985413)
(166,3.15203523635864)
(167,3.16937470436096)
(168,3.1581597328186)
(169,3.15301513671875)
(170,3.15909457206726)
(171,3.14969158172607)
(172,3.15706086158752)
(173,3.15372252464294)
(174,3.1501898765564)
(175,3.16143798828125)
(176,3.13974738121033)
(177,3.13432598114014)
(178,3.14793515205383)
(179,3.13584208488464)
(180,3.14140009880066)
(181,3.139240026474)
(182,3.13279795646668)
(183,3.13629055023193)
(184,3.14004993438721)
(185,3.14193105697632)
(186,3.12688255310059)
(187,3.13093423843384)
(188,3.13094258308411)
(189,3.11580657958984)
(190,3.12074136734009)
(191,3.13199043273926)
(192,3.11605143547058)
(193,3.12983059883118)
(194,3.12584400177)
(195,3.12263512611389)
(196,3.11649203300476)
(197,3.11256217956543)
(198,3.11427855491638)
(199,3.12432670593262)
(200,3.12764406204224)

};
\addlegendentry{GPT2-81M-LOOP Val Loss}

\addplot[dotted, red, ultra thick] coordinates {(0,3.12) (200,3.12)};
\addlegendentry{GPT2-124M Val Loss (12 Layers)}

\end{axis}
\end{tikzpicture}
\caption{Training and Validation Loss Curves for First Experiment}
\label{fig:loss_curves_first_experiment}
\end{figure}

\subsection{Second Experiment: GPT2-45M-LOOP vs. GPT2-45M}

In the comparison between the GPT2-45M-LOOP model and the GPT2-45M model, our model achieved a validation loss of 3.67 compared to 3.98, and a training loss of 3.65 compared to 3.96. By looping twice over a single transformer block, our model effectively simulates a deeper network, resulting in a substantial performance gain over the one-pass baseline. This demonstrates that even at smaller scales, our method can improve performance without increasing model size by leveraging iterative refinement.

\begin{table}[H]
\centering
\caption{Performance Metrics Emphasizing Layers and Loops for Second Experiment}
\label{tab:second_experiment}
\begin{tabular}{cccccc}
\toprule
\textbf{Model} & \textbf{Parameters} & \textbf{Layers} & \textbf{Loops} & \textbf{Train Loss} & \textbf{Validation Loss} \\
\midrule
GPT2-45M & 45M & 1 & 1 & 3.96 & 3.98 \\
GPT2-45M-LOOP & 45M & 1 & 2 & \textbf{3.65} & \textbf{3.67} \\
\bottomrule
\end{tabular}
\end{table}

\begin{figure}[H]
\centering
\begin{tikzpicture}
\begin{axis}[
    width=0.8\linewidth,
    xlabel={Training Steps},
    ylabel={Loss},
    legend pos=north east,
    grid=both,
    grid style={line width=.1pt, draw=gray!10},
    major grid style={line width=.2pt,draw=gray!50},
]

\addplot[smooth, green] plot coordinates {
(0,11.0128660202026)
(1,5.60313034057617)
(2,4.8320164680481)
(3,4.42651319503784)
(4,4.28312063217163)
(5,4.20299577713013)
(6,4.14633369445801)
(7,4.10357427597046)
(8,4.07067680358887)
(9,4.05162382125855)
(10,4.03657913208008)
(11,4.0154275894165)
(12,3.99025940895081)
(13,4.00215196609497)
(14,4.00120496749878)
(15,3.97561073303223)
(16,3.97384285926819)
(17,3.94591069221497)
(18,3.95243859291077)
(19,3.95142483711243)
(20,3.94915652275085)
(21,3.94216442108154)
(22,3.92966389656067)
(23,3.92360782623291)
(24,3.92749452590942)
(25,3.92769622802734)
(26,3.92447972297668)
(27,3.92415809631348)
(28,3.92261052131653)
(29,3.91730356216431)
(30,3.90128636360168)
(31,3.90356135368347)
(32,3.88900303840637)
(33,3.90326166152954)
(34,3.9038827419281)
(35,3.88996005058289)
(36,3.88237357139587)
(37,3.89431571960449)
(38,3.89349222183228)
(39,3.88553428649902)
(40,3.88738322257996)
(41,3.87337732315063)
(42,3.89210391044617)
(43,3.88201880455017)
(44,3.87162899971008)
(45,3.87076330184937)
(46,3.86924767494202)
(47,3.86272883415222)
(48,3.85992288589478)
(49,3.8757050037384)
(50,3.8606493473053)
(51,3.87962770462036)
(52,3.87682700157166)
(53,3.87487173080444)
(54,3.86714243888855)
(55,3.86138010025024)
(56,3.86437010765076)
(57,3.85700368881226)
(58,3.84049868583679)
(59,3.84281849861145)
(60,3.84869456291199)
(61,3.85425233840942)
(62,3.83694696426392)
(63,3.84487390518188)
(64,3.84338164329529)
(65,3.84075808525085)
(66,3.84959077835083)
(67,3.84846615791321)
(68,3.84059238433838)
(69,3.83209109306335)
(70,3.83423948287964)
(71,3.8467230796814)
(72,3.8451418876648)
(73,3.83391642570496)
(74,3.82897210121155)
(75,3.83515501022339)
(76,3.8320996761322)
(77,3.82348394393921)
(78,3.82064747810364)
(79,3.80911612510681)
(80,3.81763935089111)
(81,3.80470585823059)
(82,3.82579326629639)
(83,3.81546783447266)
(84,3.82643818855286)
(85,3.81283140182495)
(86,3.82427430152893)
(87,3.80877828598022)
(88,3.80804014205933)
(89,3.80902504920959)
(90,3.79793739318848)
(91,3.82200622558594)
(92,3.80846810340881)
(93,3.83217835426331)
(94,3.80144858360291)
(95,3.80735659599304)
(96,3.79683542251587)
(97,3.81022310256958)
(98,3.78682398796082)
(99,3.79240536689758)
(100,3.79445934295654)
(101,3.78347015380859)
(102,3.78699254989624)
(103,3.78850483894348)
(104,3.78750586509705)
(105,3.78882789611816)
(106,3.77648782730103)
(107,3.77265620231628)
(108,3.77626991271973)
(109,3.7883563041687)
(110,3.78928780555725)
(111,3.76937985420227)
(112,3.76696801185608)
(113,3.76565742492676)
(114,3.78660345077515)
(115,3.75475406646729)
(116,3.76303124427795)
(117,3.76113271713257)
(118,3.78529787063599)
(119,3.76909947395325)
(120,3.77850270271301)
(121,3.77145910263062)
(122,3.76722311973572)
(123,3.76154541969299)
(124,3.77107262611389)
(125,3.7446870803833)
(126,3.75294351577759)
(127,3.76567935943604)
(128,3.75391530990601)
(129,3.7400815486908)
(130,3.74878883361816)
(131,3.76273655891418)
(132,3.73215556144714)
(133,3.74061131477356)
(134,3.74632930755615)
(135,3.75290131568909)
(136,3.73061227798462)
(137,3.74390935897827)
(138,3.74995517730713)
(139,3.72989535331726)
(140,3.73601675033569)
(141,3.7241702079773)
(142,3.71154761314392)
(143,3.71500682830811)
(144,3.73949146270752)
(145,3.71462941169739)
(146,3.71493744850159)
(147,3.71601891517639)
(148,3.72761344909668)
(149,3.70993232727051)
(150,3.71623849868774)
(151,3.72764444351196)
(152,3.73779010772705)
(153,3.71255564689636)
(154,3.73581218719482)
(155,3.70369243621826)
(156,3.70510339736938)
(157,3.70946168899536)
(158,3.68957448005676)
(159,3.71803593635559)
(160,3.71233224868774)
(161,3.70628809928894)
(162,3.68872094154358)
(163,3.68873739242554)
(164,3.68924999237061)
(165,3.6995587348938)
(166,3.67573404312134)
(167,3.69422245025635)
(168,3.68344521522522)
(169,3.69141912460327)
(170,3.68312859535217)
(171,3.67330503463745)
(172,3.68294215202332)
(173,3.68732333183289)
(174,3.67558908462524)
(175,3.67636966705322)
(176,3.67763543128967)
(177,3.67863702774048)
(178,3.67754769325256)
(179,3.66741394996643)
(180,3.68449759483337)
(181,3.67443633079529)
(182,3.67532157897949)
(183,3.66336894035339)
(184,3.67067623138428)
(185,3.67045903205872)
(186,3.66720652580261)
(187,3.66521000862122)
(188,3.66514706611633)
(189,3.66908383369446)
(190,3.67208981513977)
(191,3.65736031532288)
(192,3.67763233184814)
(193,3.6644082069397)
(194,3.66249752044678)
(195,3.65112090110779)
(196,3.64909887313843)
(197,3.67131876945496)
(198,3.67189788818359)
(199,3.65144634246826)
(200,3.67333078384399)

};
\addlegendentry{GPT2-45M-LOOP (2 Loops) Train Loss}

\addplot[smooth, blue, dashed] plot coordinates {
(0,11.0137662887573)
(1,5.60678577423096)
(2,4.83149671554565)
(3,4.43286848068237)
(4,4.30172920227051)
(5,4.20873689651489)
(6,4.1559157371521)
(7,4.10591602325439)
(8,4.0798716545105)
(9,4.07264852523804)
(10,4.05024480819702)
(11,4.0294075012207)
(12,4.01427602767944)
(13,3.99729824066162)
(14,3.99737524986267)
(15,3.98920345306397)
(16,3.96545553207397)
(17,3.95313119888306)
(18,3.96178793907166)
(19,3.96047854423523)
(20,3.9480938911438)
(21,3.94352912902832)
(22,3.93618035316467)
(23,3.94126582145691)
(24,3.92749667167664)
(25,3.91679835319519)
(26,3.93967461585999)
(27,3.91472840309143)
(28,3.93264961242676)
(29,3.91857933998108)
(30,3.91547846794128)
(31,3.90373229980469)
(32,3.92689919471741)
(33,3.9120819568634)
(34,3.88708400726318)
(35,3.90831851959229)
(36,3.90032052993774)
(37,3.89461421966553)
(38,3.89904165267944)
(39,3.89171600341797)
(40,3.8990261554718)
(41,3.89882040023804)
(42,3.88241147994995)
(43,3.88250184059143)
(44,3.88773488998413)
(45,3.88120722770691)
(46,3.87686800956726)
(47,3.87805867195129)
(48,3.88237404823303)
(49,3.86885619163513)
(50,3.88742136955261)
(51,3.8818678855896)
(52,3.87857723236084)
(53,3.87716817855835)
(54,3.85637331008911)
(55,3.86798286437988)
(56,3.8756115436554)
(57,3.87229776382446)
(58,3.86389827728272)
(59,3.84838223457336)
(60,3.88451862335205)
(61,3.86708045005798)
(62,3.85824346542358)
(63,3.87730169296265)
(64,3.8487982749939)
(65,3.84597706794739)
(66,3.84719514846802)
(67,3.85603547096252)
(68,3.84293484687805)
(69,3.83781242370605)
(70,3.83868861198425)
(71,3.82573175430298)
(72,3.85093712806702)
(73,3.83835911750793)
(74,3.82650661468506)
(75,3.84211301803589)
(76,3.82158660888672)
(77,3.84807920455933)
(78,3.83478760719299)
(79,3.83588409423828)
(80,3.8429217338562)
(81,3.82728338241577)
(82,3.82960510253906)
(83,3.8392276763916)
(84,3.83497619628906)
(85,3.83377981185913)
(86,3.8292396068573)
(87,3.83377742767334)
(88,3.81947731971741)
(89,3.82442021369934)
(90,3.82025170326233)
(91,3.81881785392761)
(92,3.80707120895386)
(93,3.82188320159912)
(94,3.81194877624512)
(95,3.81562900543213)
(96,3.80391049385071)
(97,3.78977870941162)
(98,3.8074517250061)
(99,3.81271839141846)
(100,3.80310702323914)
(101,3.80253911018372)
(102,3.79014658927918)
(103,3.79282474517822)
(104,3.79706501960754)
(105,3.7915678024292)
(106,3.80909633636475)
(107,3.79579782485962)
(108,3.7885320186615)
(109,3.80870079994202)
(110,3.78250408172607)
(111,3.78332734107971)
(112,3.79709553718567)
(113,3.78649377822876)
(114,3.77305054664612)
(115,3.77471017837524)
(116,3.78023600578308)
(117,3.77657318115234)
(118,3.78172612190247)
(119,3.77776956558228)
(120,3.77270984649658)
(121,3.76917505264282)
(122,3.78344035148621)
(123,3.77148985862732)
(124,3.77404093742371)
(125,3.7716338634491)
(126,3.75346565246582)
(127,3.75314211845398)
(128,3.75621056556702)
(129,3.75424599647522)
(130,3.75210332870483)
(131,3.76477909088135)
(132,3.73897838592529)
(133,3.75814366340637)
(134,3.74630641937256)
(135,3.76898431777954)
(136,3.75169134140015)
(137,3.74484872817993)
(138,3.75252938270569)
(139,3.72884798049927)
(140,3.73760151863098)
(141,3.74644470214844)
(142,3.75082206726074)
(143,3.73171806335449)
(144,3.73738861083984)
(145,3.73675966262817)
(146,3.73123526573181)
(147,3.73404836654663)
(148,3.73925876617432)
(149,3.73668217658997)
(150,3.71467304229736)
(151,3.72514581680298)
(152,3.71442556381226)
(153,3.72105979919434)
(154,3.70718836784363)
(155,3.7238290309906)
(156,3.70708656311035)
(157,3.71472358703613)
(158,3.71473670005798)
(159,3.71255278587341)
(160,3.71256470680237)
(161,3.7107834815979)
(162,3.70712304115295)
(163,3.69525218009949)
(164,3.69029664993286)
(165,3.71318936347961)
(166,3.71631908416748)
(167,3.702472448349)
(168,3.69245362281799)
(169,3.6885507106781)
(170,3.69248104095459)
(171,3.6976580619812)
(172,3.67786407470703)
(173,3.69330239295959)
(174,3.68493747711182)
(175,3.6821506023407)
(176,3.68862390518188)
(177,3.69872522354126)
(178,3.6935760974884)
(179,3.68367123603821)
(180,3.68488073348999)
(181,3.68787145614624)
(182,3.67829561233521)
(183,3.68835759162903)
(184,3.66853427886963)
(185,3.67044496536255)
(186,3.67679882049561)
(187,3.67459177970886)
(188,3.68067622184753)
(189,3.68503642082214)
(190,3.68856739997864)
(191,3.67102360725403)
(192,3.68354797363281)
(193,3.67031669616699)
(194,3.67117929458618)
(195,3.6698272228241)
(196,3.67034912109375)
(197,3.66530203819275)
(198,3.67920804023743)
(199,3.67244744300842)
(200,3.68048524856567)

};
\addlegendentry{GPT2-45M-LOOP Val Loss}

\addplot[smooth, orange] plot coordinates {
(0,11.0268602371216)
(1,5.5972752571106)
(2,4.85482597351074)
(3,4.60941362380981)
(4,4.51668500900269)
(5,4.43684005737305)
(6,4.40167474746704)
(7,4.36152505874634)
(8,4.34077310562134)
(9,4.33282804489136)
(10,4.31550884246826)
(11,4.29143142700195)
(12,4.28508424758911)
(13,4.27399492263794)
(14,4.2675666809082)
(15,4.26377058029175)
(16,4.24666547775269)
(17,4.22215509414673)
(18,4.24802160263062)
(19,4.24753856658936)
(20,4.22933340072632)
(21,4.23036623001099)
(22,4.21314239501953)
(23,4.2251558303833)
(24,4.20900774002075)
(25,4.2019157409668)
(26,4.22616004943848)
(27,4.20052337646484)
(28,4.21939277648926)
(29,4.20206308364868)
(30,4.20164299011231)
(31,4.20389223098755)
(32,4.213942527771)
(33,4.20782279968262)
(34,4.18440532684326)
(35,4.20163631439209)
(36,4.18925476074219)
(37,4.18055248260498)
(38,4.18367385864258)
(39,4.18471384048462)
(40,4.18538618087769)
(41,4.18186855316162)
(42,4.17189311981201)
(43,4.17347431182861)
(44,4.18056297302246)
(45,4.18229484558106)
(46,4.16809463500977)
(47,4.17324209213257)
(48,4.17658424377441)
(49,4.16035938262939)
(50,4.17762422561646)
(51,4.16743612289429)
(52,4.16930294036865)
(53,4.16965007781982)
(54,4.14899396896362)
(55,4.15931749343872)
(56,4.16251230239868)
(57,4.16364622116089)
(58,4.15943670272827)
(59,4.14364576339722)
(60,4.17685794830322)
(61,4.14922714233398)
(62,4.15813493728638)
(63,4.17024421691895)
(64,4.1441593170166)
(65,4.13543462753296)
(66,4.14275407791138)
(67,4.15120077133179)
(68,4.14096021652222)
(69,4.12702226638794)
(70,4.12892436981201)
(71,4.12803220748901)
(72,4.14573526382446)
(73,4.14396953582764)
(74,4.12806606292725)
(75,4.1471905708313)
(76,4.11953496932983)
(77,4.15101909637451)
(78,4.13506555557251)
(79,4.13545370101929)
(80,4.14093446731567)
(81,4.1252613067627)
(82,4.13412094116211)
(83,4.13021993637085)
(84,4.13424968719482)
(85,4.12301063537598)
(86,4.12809896469116)
(87,4.12818098068237)
(88,4.11929130554199)
(89,4.1183819770813)
(90,4.11221504211426)
(91,4.12424182891846)
(92,4.10742616653442)
(93,4.11919832229614)
(94,4.10395574569702)
(95,4.10946369171143)
(96,4.1059775352478)
(97,4.10034370422363)
(98,4.1099796295166)
(99,4.11551713943481)
(100,4.10102224349976)
(101,4.09711790084839)
(102,4.09471035003662)
(103,4.08714771270752)
(104,4.10059642791748)
(105,4.0930495262146)
(106,4.11018800735474)
(107,4.09575843811035)
(108,4.0911979675293)
(109,4.10248756408691)
(110,4.0900149345398)
(111,4.08180379867554)
(112,4.08862018585205)
(113,4.08489751815796)
(114,4.07028770446777)
(115,4.07179737091064)
(116,4.07810878753662)
(117,4.08358383178711)
(118,4.07052755355835)
(119,4.07197856903076)
(120,4.07010793685913)
(121,4.06698799133301)
(122,4.08735466003418)
(123,4.06795501708984)
(124,4.07217121124268)
(125,4.06763648986816)
(126,4.05369663238525)
(127,4.05719709396362)
(128,4.05888175964356)
(129,4.05832052230835)
(130,4.05400466918945)
(131,4.06120491027832)
(132,4.04089546203613)
(133,4.05766105651856)
(134,4.04729175567627)
(135,4.06520318984985)
(136,4.04696273803711)
(137,4.04267597198486)
(138,4.05049276351929)
(139,4.02992343902588)
(140,4.04306268692017)
(141,4.04365873336792)
(142,4.05517530441284)
(143,4.03678894042969)
(144,4.03758144378662)
(145,4.03802061080933)
(146,4.03395223617554)
(147,4.03899192810059)
(148,4.03484439849854)
(149,4.04023790359497)
(150,4.02243614196777)
(151,4.02123355865479)
(152,4.01106214523315)
(153,4.02505683898926)
(154,4.01172208786011)
(155,4.02141427993774)
(156,4.0122561454773)
(157,4.01438426971436)
(158,4.0182032585144)
(159,4.01308441162109)
(160,4.01164102554321)
(161,4.01079225540161)
(162,4.01328659057617)
(163,4.00207376480103)
(164,4.00216436386108)
(165,4.01498746871948)
(166,4.0199031829834)
(167,4.00365591049194)
(168,4.00235652923584)
(169,3.99342703819275)
(170,4.00113821029663)
(171,3.99926567077637)
(172,3.98198676109314)
(173,3.99771785736084)
(174,3.98955225944519)
(175,3.99163031578064)
(176,3.98543238639832)
(177,3.99648571014404)
(178,3.99441075325012)
(179,3.9862813949585)
(180,3.98860144615173)
(181,3.99327230453491)
(182,3.98623299598694)
(183,3.99432134628296)
(184,3.97626709938049)
(185,3.97453117370605)
(186,3.98264718055725)
(187,3.97662711143494)
(188,3.99023509025574)
(189,3.98035764694214)
(190,3.98908805847168)
(191,3.98196291923523)
(192,3.98534822463989)
(193,3.98017525672913)
(194,3.97644472122192)
(195,3.98114490509033)
(196,3.97657895088196)
(197,3.9733099937439)
(198,3.98914909362793)
(199,3.97895193099976)
(200,3.9875168800354)

};
\addlegendentry{GPT2-45M Train Loss}

\addplot[smooth, red, dashed] plot coordinates {
(0,11.0268602371216)
(1,5.5972752571106)
(2,4.85482597351074)
(3,4.60941362380981)
(4,4.51668500900269)
(5,4.43684005737305)
(6,4.40167474746704)
(7,4.36152505874634)
(8,4.34077310562134)
(9,4.33282804489136)
(10,4.31550884246826)
(11,4.29143142700195)
(12,4.28508424758911)
(13,4.27399492263794)
(14,4.2675666809082)
(15,4.26377058029175)
(16,4.24666547775269)
(17,4.22215509414673)
(18,4.24802160263062)
(19,4.24753856658936)
(20,4.22933340072632)
(21,4.23036623001099)
(22,4.21314239501953)
(23,4.2251558303833)
(24,4.20900774002075)
(25,4.2019157409668)
(26,4.22616004943848)
(27,4.20052337646484)
(28,4.21939277648926)
(29,4.20206308364868)
(30,4.20164299011231)
(31,4.20389223098755)
(32,4.213942527771)
(33,4.20782279968262)
(34,4.18440532684326)
(35,4.20163631439209)
(36,4.18925476074219)
(37,4.18055248260498)
(38,4.18367385864258)
(39,4.18471384048462)
(40,4.18538618087769)
(41,4.18186855316162)
(42,4.17189311981201)
(43,4.17347431182861)
(44,4.18056297302246)
(45,4.18229484558106)
(46,4.16809463500977)
(47,4.17324209213257)
(48,4.17658424377441)
(49,4.16035938262939)
(50,4.17762422561646)
(51,4.16743612289429)
(52,4.16930294036865)
(53,4.16965007781982)
(54,4.14899396896362)
(55,4.15931749343872)
(56,4.16251230239868)
(57,4.16364622116089)
(58,4.15943670272827)
(59,4.14364576339722)
(60,4.17685794830322)
(61,4.14922714233398)
(62,4.15813493728638)
(63,4.17024421691895)
(64,4.1441593170166)
(65,4.13543462753296)
(66,4.14275407791138)
(67,4.15120077133179)
(68,4.14096021652222)
(69,4.12702226638794)
(70,4.12892436981201)
(71,4.12803220748901)
(72,4.14573526382446)
(73,4.14396953582764)
(74,4.12806606292725)
(75,4.1471905708313)
(76,4.11953496932983)
(77,4.15101909637451)
(78,4.13506555557251)
(79,4.13545370101929)
(80,4.14093446731567)
(81,4.1252613067627)
(82,4.13412094116211)
(83,4.13021993637085)
(84,4.13424968719482)
(85,4.12301063537598)
(86,4.12809896469116)
(87,4.12818098068237)
(88,4.11929130554199)
(89,4.1183819770813)
(90,4.11221504211426)
(91,4.12424182891846)
(92,4.10742616653442)
(93,4.11919832229614)
(94,4.10395574569702)
(95,4.10946369171143)
(96,4.1059775352478)
(97,4.10034370422363)
(98,4.1099796295166)
(99,4.11551713943481)
(100,4.10102224349976)
(101,4.09711790084839)
(102,4.09471035003662)
(103,4.08714771270752)
(104,4.10059642791748)
(105,4.0930495262146)
(106,4.11018800735474)
(107,4.09575843811035)
(108,4.0911979675293)
(109,4.10248756408691)
(110,4.0900149345398)
(111,4.08180379867554)
(112,4.08862018585205)
(113,4.08489751815796)
(114,4.07028770446777)
(115,4.07179737091064)
(116,4.07810878753662)
(117,4.08358383178711)
(118,4.07052755355835)
(119,4.07197856903076)
(120,4.07010793685913)
(121,4.06698799133301)
(122,4.08735466003418)
(123,4.06795501708984)
(124,4.07217121124268)
(125,4.06763648986816)
(126,4.05369663238525)
(127,4.05719709396362)
(128,4.05888175964356)
(129,4.05832052230835)
(130,4.05400466918945)
(131,4.06120491027832)
(132,4.04089546203613)
(133,4.05766105651856)
(134,4.04729175567627)
(135,4.06520318984985)
(136,4.04696273803711)
(137,4.04267597198486)
(138,4.05049276351929)
(139,4.02992343902588)
(140,4.04306268692017)
(141,4.04365873336792)
(142,4.05517530441284)
(143,4.03678894042969)
(144,4.03758144378662)
(145,4.03802061080933)
(146,4.03395223617554)
(147,4.03899192810059)
(148,4.03484439849854)
(149,4.04023790359497)
(150,4.02243614196777)
(151,4.02123355865479)
(152,4.01106214523315)
(153,4.02505683898926)
(154,4.01172208786011)
(155,4.02141427993774)
(156,4.0122561454773)
(157,4.01438426971436)
(158,4.0182032585144)
(159,4.01308441162109)
(160,4.01164102554321)
(161,4.01079225540161)
(162,4.01328659057617)
(163,4.00207376480103)
(164,4.00216436386108)
(165,4.01498746871948)
(166,4.0199031829834)
(167,4.00365591049194)
(168,4.00235652923584)
(169,3.99342703819275)
(170,4.00113821029663)
(171,3.99926567077637)
(172,3.98198676109314)
(173,3.99771785736084)
(174,3.98955225944519)
(175,3.99163031578064)
(176,3.98543238639832)
(177,3.99648571014404)
(178,3.99441075325012)
(179,3.9862813949585)
(180,3.98860144615173)
(181,3.99327230453491)
(182,3.98623299598694)
(183,3.99432134628296)
(184,3.97626709938049)
(185,3.97453117370605)
(186,3.98264718055725)
(187,3.97662711143494)
(188,3.99023509025574)
(189,3.98035764694214)
(190,3.98908805847168)
(191,3.98196291923523)
(192,3.98534822463989)
(193,3.98017525672913)
(194,3.97644472122192)
(195,3.98114490509033)
(196,3.97657895088196)
(197,3.9733099937439)
(198,3.98914909362793)
(199,3.97895193099976)
(200,3.9875168800354)

};
\addlegendentry{GPT2-45M Val Loss}

\end{axis}
\end{tikzpicture}
\caption{Training and Validation Loss Curves for Second Experiment}
\label{fig:loss_curves_second_experiment}
\end{figure}
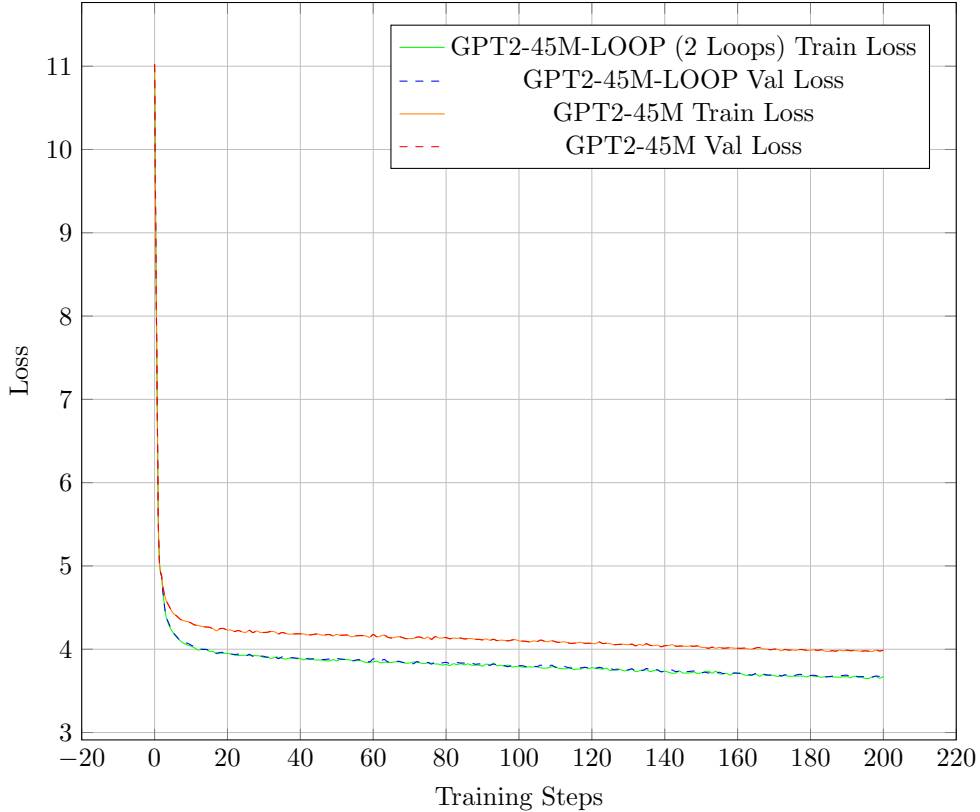

\subsection{Discussion}

The results indicate that our loop architecture effectively leverages iterative refinement to enhance model performance without increasing the parameter count. In the first experiment, despite having significantly fewer parameters and layers compared to the GPT2-124M baseline, our GPT2-81M-LOOP and GPT2-67M-LOOP models achieve comparable validation losses. This suggests that the iterative loops over the transformer blocks allow the models to capture complex patterns and dependencies that would otherwise require additional layers or parameters.

In the second experiment, the GPT2-45M-LOOP model significantly outperforms the GPT2-45M baseline. By looping twice over a single transformer block, our model effectively simulates a deeper network, resulting in a reduction of the validation loss from 3.98 to 3.67. This demonstrates that even at smaller scales, iterative refinement can compensate for a limited number of layers, enhancing the model's expressiveness.

The increased training epoch times for our loop neural networks are relatively modest compared to the performance gains. For instance, the training time for the GPT2-45M-LOOP model is only 18\% longer than the GPT2-45M baseline, while achieving a substantial improvement in validation loss. This indicates that the computational overhead introduced by the loops is manageable, making our approach practical for applications where longer inference times are acceptable in exchange for improved performance.

Our experiments highlight the advantage of our setup in terms of resource efficiency and scalability. By avoiding the need for additional parameters or larger models, our loop neural network is more accessible for researchers and practitioners with limited computational resources. Furthermore, since our approach does not rely on additional training data or specialized datasets, it offers a scalable solution for enhancing model performance in various language modeling tasks.

\section{Conclusion}

We have presented the loop neural network, a novel approach that enables smaller neural network models to achieve better results on lower-end devices by leveraging longer inference times through iterative refinement. By iteratively refining the hidden state through residual connections over multiple loops of a model subset, our method captures complex patterns and dependencies more effectively than conventional one-pass models. Our experiments demonstrate that the loop neural networks can achieve improved performance over baseline models of the same size and comparable performance to larger models with fewer parameters. Significantly, these performance improvements are achieved without the need for additional training data, highlighting the efficiency and accessibility of our approach. This work opens up new possibilities for neural network architectures, particularly for tasks that benefit from deeper computational reasoning on resource-constrained devices.

\bibliographystyle{apalike}
\bibliography{references}

\end{document}